\def\year{2020}\relax
\documentclass[letterpaper]{article} 
\usepackage{aaai20}  
\usepackage{times}  
\usepackage{helvet} 
\usepackage{courier}  
\usepackage[hyphens]{url}  
\usepackage{graphicx} 
\usepackage{graphicx}  
\usepackage[utf8]{inputenc} 
\usepackage{booktabs}       
\usepackage{microtype}      

\usepackage[ruled, noline]{algorithm2e}
\usepackage{graphicx}
\usepackage{amsmath}
\usepackage{amssymb}
\usepackage{subfigure}

\def\eq#1{Eq. (\ref{#1})}

\def\fig#1{Figure \ref{#1}}
\def\tabl#1{Table \ref{#1}}
\def\alg#1{Algorithm~\ref{#1}}
\def\se#1{Section~\ref{#1}}

\def\th{\mbox{\em th} }

\newcommand{\eg}{{\em e.g.}}
\newcommand{\ie}{{\em i.e.}}
\newcommand{\etc}{{\em etc.}}
\newcommand{\etal}{ {\em et al.}}

\newcommand{\swapped}[2]{
{\begin{tabular}{cc}
\setlength{\tabcolsep}{1pt}
\rotatebox[origin=c]{90}{{\scriptsize Fixed Content}} & \begin{tabular}{c}
{\scriptsize Fixed Style}\\
\includegraphics[width=#1]{#2}
\end{tabular}
\end{tabular}}
}

\SetKwInOut{Input}{Input}
\SetKwInOut{Output}{Output}
\SetKwFor{RepTimes}{repeat}{times}{end}

\title{Adversarial Disentanglement with Grouped Observations}
\author{
Jozsef Nemeth\\ 
Ultinous, Hungary\\
jnemeth@ultinous.com 
}

\begin{document}

\maketitle

\begin{abstract}
We consider the disentanglement of the representations of the relevant attributes of the data ({\it content}) from all other factors of variations ({\it style}) using Variational Autoencoders. Some recent works addressed this problem by utilizing grouped observations, where the content attributes are assumed to be common within each group, while there is no any supervised information on the style factors. In many cases, however, these methods fail to prevent the models from using the style variables to encode content related features as well. This work supplements these algorithms with a method that eliminates the content information in the style representations. For that purpose the training objective is augmented to minimize an appropriately defined mutual information term in an adversarial way. Experimental results and comparisons on image datasets show that the resulting method can efficiently separate the content and style related attributes and generalizes to unseen data. 
\end{abstract}

\section{Introduction}
In the field of representation learning~\cite{Bengio2013}, autoencoder based approaches~\cite{Tschannen2018} are among the most effective methods to learn compact and meaningful representations even without any supervision. Such representations then can be used to solve downstream tasks like classification or clustering efficiently. Variational Autoencoders (VAEs)~\cite{Kingma2014} attracted probably the most attention in recent years. By employing stochastic variational inference~\cite{Zhang2018}, VAEs can learn intractable posterior distributions of the latent variables.

Disentangled representation learning~\cite{Desjardins2012,Kumar2018,Achille2017,Esmaeili2019,Xiang2019} aims to assign the different factors of variations to different dimensions of the representation vectors. This problem has been heavily studied in recent years, for example, it has been proved that learning disentangled representations without any inductive bias or supervision is  theoretically impossible~\cite{Locatello2019}. At the same time, it has been observed that using increased regularization weight, the VAE model learns disentangled latent spaces~\cite{Higgins2017}. 

In this work, we consider the problem of learning a disentangled representation in which the relevant attributes of the observations (which determine some kind of identity or class of a given sample) and the other irrelevant factors (like pose, lighting conditions, \etc) are separated. Following the notations of \cite{Bouchacourt2018}, we refer to the relevant attributes and the other factors as {\it content} and {\it style}, respectively. Recently, grouped observations based methods have been proposed for content-style disentanglement~\cite{Bouchacourt2018,Hosoya2019}. Grouping data elements that represent the same content is a way to induce weak supervision, and the main motivation behind this approach is that collecting and cleaning such datasets is less expensive. For example, in case of classification based methods it is expected that different classes represent different identities. At the same time, in case of learning from grouped data, we only expect that members of a given group share the same content, whereas different groups are not required to represent different classes. These methods~\cite{Bouchacourt2018,Hosoya2019} share the idea of accumulating the content representation of the individual group members into a single common group-level content representation, while they learn individual style representations for each of the group members. However, these methods have no explicit mechanism to prevent the model from using the style variable to represent not only style attributes but also certain amount of content information. For example, the number of dimensions for the style variable had to be adjusted for the style variability, otherwise the model used the style variable to learn content related aspects of the data~\cite{Hosoya2019}. Our experiments also confirm this observation.

We show that the results of these methods can be significantly improved by suppressing the content information in the style variable. More specifically, the proposed method trains a neural network to estimate the mutual information~\cite{Belghazi2018} between the style representations and the observations corresponding to the same groups whereas the encoder network is trained in an adversarial manner. Experiments prove that this enhanced training process provides more useful representations. The improvements are the most significant for small group sizes, which makes the proposed method more applicable in practice.

This paper is organized as follows. First, we briefly overview the related literature and the grouped observations based methods~\cite{Bouchacourt2018,Hosoya2019} in \se{se:relatedWork}. Then \se{se:adversarialDisentanglement} introduces our adversarial disentanglement approach. Experimental results and comparisons can be found in \se{se:experiments} and in the supplementary material. The source code of the experiments and supplementary material are available online\footnote{https://github.com/jonemeth/aaai20}.

\section{Related Work}
\label{se:relatedWork}
The different variants of autoencoders are powerful tools for learning efficient data representations~\cite{Tschannen2018}. Among them Variational Autoencoders (VAEs)~\cite{Kingma2014} are the most heavily studied approaches. VAEs were developed to model probabilistic latent representations based on the variational inference principle. It has also been shown that by slightly modifying the VAEs objective function, the models tend to disentangle the latent space~\cite{Higgins2017,Burgess2017}. This phenomenon was further investigated by~\citeauthor{Kim2018}~(\citeyear{Kim2018}) leading to the FactorVAE method. Chen~\etal~(\citeyear{RickyChen2018}) also studied the disentangling properties of VAEs. They proposed to split the regularization term up to four parts and assign different weights to each term. Another group of methods uses structured latent spaces to learn efficient data representations~\cite{Grathwohl2016,Maaloe2019,Balcan2016,Klys2018}. It has also been demonstrated that VAEs are capable to separate the continuous and discrete generative factors~\cite{Dupont2018}, furthermore, disentanglement can also be achieved in the semi-supervised setting~\cite{Kingma2014b,Siddharth2017}.

Generative Adversarial Networks~\cite{Goodfellow2014} are probabilistic methods designed to model high dimensional data distributions. They can be extended to representation learning, for example, the InfoGAN model~\cite{Chen2016} can learn disentangled representations by maximizing the mutual information between the generated data and a subset of latent variables. Another GAN based model has been proposed recently which can be used to control the content and the style of the generated images~\cite{Chen2018}. During training, the generator aims to produce image pairs from shared content vectors but independent style vectors, while the discriminator tries to distinguish these fake pairs from real image pairs. The generator can compete with the discriminator only by producing image pairs containing the same object. Donahue~\etal~(\citeyear{Donahue2018}) proposed a similar method although their approach to the discriminator differs.

Adversarial training to disentangle content and style attributes has been proposed recently~\cite{Mathieu2016}. The basic idea of this work is to pair the content variable of a data sample with the style variable of another sample from the same group during training. The resulting representation should be suitable to recover the the latter sample using the decoder. This approach is similar in spirit with the grouped observations based methods~\cite{Bouchacourt2018,Hosoya2019}, on which the current work is based. The main difference is that the latter methods do not require access to class labels. Kulkarni~\etal (\citeyear{Kulkarni2015}) proposed a method to disentangle the generative factors and create a compact and interpretable representation. The algorithm encourages the different dimensions of the latent variable to represent specific attributes of the images. Recently, Wu~\etal~(\citeyear{Wu2019}) proposed an architecture to disentangle geometry and style information using prior knowledge on structure. Other GAN and autoencoder based generative models were successfully applied to face frontalization~\cite{Yin2017} and recognition~\cite{Liu2018}. Learning representations that are invariant to specific factors is also highly related to the current work~\cite{Jaiswal2018}. It has recently been shown that invariance can be obtained without adversarial training as well~\cite{Moyer2018}.

\subsection{Multi-Level Variational Autoencoders}
\label{se:groupedBased}
To separate the content and style representations~\citeauthor{Bouchacourt2018}~(\citeyear{Bouchacourt2018}) introduced the Multi-Level Variational Autoencoder (MLVAE) which utilizes a dataset of grouped observations. In this section, we briefly overview this approach. Let the training dataset consist of $N$ observation groups ${\bf x}^n=\left\{x^n_1,\dots, x^n_{K_n}\right\}$ for $n=1,\dots,N$, where $K_n=|{\bf x}^n|$ denote the number of individual observations within the group and $x^n_i\in \mathbb{R}^d$ is the $i$\th member of the $n$\th group. The group index $n$ is omitted sometimes for the sake of simplicity and in this work we consider only datasets with $K_n=K$,~\ie, all the groups in the dataset are of the same size. The underlying empirical data distribution represented by the dataset is denoted by $p_{\mathcal D}( {\bf x} )$. We also define the distribution of individual observations $p_{\mathcal D}(x)$. Note that sampling from $p_{\mathcal D}(x)$ can be performed by first sampling a group from  $p_{\mathcal D}( {\bf x} )$ then choosing a member uniformly at random. We assume that the members in a given group share some content attributes that we want to capture by the latent variable $c\in \mathbb{R}^{d_c}$, while the individual group members have style attributes represented by the latent variables ${\bf s} = \left\{s_i\in \mathbb{R}^{d_s}:i=1,\dots,K\right\}$. The Evidence Lower Bound (ELBO) for a group~\cite{Bouchacourt2018} is given by:
\begin{align}
\log p_\theta({\bf x}) 
&\geq \mathbb{E}_{q_\phi(c, {\bf s} | {\bf x})}  \sum_{i=1}^K\log p_\theta(x_i | c, s_i )  \nonumber\\
&-\sum_{i=1}^K KL(q_\phi(s_i|x_i)|p(s_i))\nonumber\\
&- KL(q_\phi(c|{\bf x})|p(c)) ={\mathcal L}_{\theta,\phi}({\bf x}),
\label{eq:groupELBO}
\end{align}
where $p(s_i)$ and $p(c)$ are priors on the latent variables, while $\theta$ is the parameter of the generative model and $\phi$ is the variational parameter of both the content and style variables. The training process maximizes this lower bound for the whole dataset:
\begin{equation}
{\mathcal L}_{\theta,\phi}({\mathcal D}) = \frac{1}{N}\sum_{n=1}^N  {\mathcal L}_{\theta,\phi}({\bf x}^n).
\label{eq:fullELBO}
\end{equation}
As usually in case of Variational Autoencoders, both the generative model $p_\theta$ and the approximate inference model $q_\phi$ are realized by neural networks $P_\theta$ and $Q_\phi$ with parameters $\theta$ and $\phi$, respectively. In the experiments presented in this paper the encoder neural network estimated the expected value and log-variance of the Gaussian approximate posterior distribution and we used standard normal priors everywhere. Furthermore, we considered Bernoulli likelihood for binary images and Gaussian likelihood (with fixed variance of $1.0$) for color images.

\subsubsection*{Content Accumulation}
For the MLVAE method, it was proposed to define the single content posterior for a given group based on the content encodings obtained from the individual observations withing that group. More specifically, the product of the posterior density functions provided by the encoder was considered:
\begin{equation}
q_\phi(c|{\bf x}) = \frac{1}{Z}\prod_{i=1}^K q_\phi(c|x_i),
\label{eq:accByMul}
\end{equation}

where $Z$ is a normalization constant. Although theoretically other approaches are also feasible (such as accumulating as a Gaussian mixture), in our experiments we focus on ~\eq{eq:accByMul}. In \cite{Hosoya2019}, another accumulation method has been proposed. Our experiments in the supplemental show that their simplified accumulation approach slightly improves the results, but the adversarial method described in the next section is still important to obtain significantly better disentanglement.

\section{Adversarial Disentanglement}
\label{se:adversarialDisentanglement}
From the point of view of disentangling the latent space, the accumulation of the content posteriors over group members effectively discourages the encoder from storing style related information in the content variables. This is because the decoder reconstructs the inputs from the accumulated contents and individual styles. On the other hand, there is no explicit mechanism that would prevent the model from encoding content in the style variable $s$. For example, in case of the Chairs dataset~\cite{Yang2015}, in which the only factors of style variations are two rotation angles, it is important to choose a low number of dimensions for $s$ (\eg, $d_s=2$) to prevent the model from storing content in it~\cite{Hosoya2019}.

Here we argue that the reason why maximizing the Group-ELBO in \eq{eq:groupELBO} can result in disentanglement (at least to some extent) is the imbalanced regularization approach. Informally speaking, \eq{eq:groupELBO} puts more weight on regularizing the style variable (which in turn minimizes its information content about the data~\cite{Kim2018}) and thus encourages the encoder to use the content variable as well. As experimental results in \se{se:experiments} and in the supplementary material show, larger group size leads to better disentanglement, however, for small group sizes the algorithm ignores the content variable $c$, and uses only $s$ to represent all aspects of the data. One might expect to obtain better results by increasing the regularization on $s$ and thus encouraging the model to use the content vector $c$. Experiments presented in \se{se:experiments} show that while this approach increases the disentanglement indeed, it also harms the overall performance of the method. This is because increased regularization reduces the capacity of $s$ and weaker style encoding affects the quality of the content representation as well.

In this section, we show how to minimize only the content related information in the style representation. The group members are assumed to share the same content, but the style attributes are independent. Therefore, we expect that the style representation of a given sample can not be inferred from other data samples within the same group. The content information encoded in $s$ can be measured as the mutual information between $s$ and other data samples in the given group. To formulate this concept, let us first define:
\begin{equation}
r_{\phi}(s|x) = r_{\phi}(s|x=x^n_i) \equiv \frac{1}{K-1}\sum_{j=1, j\neq i}^K q_\phi(s| x^n_j)
\label{eq:r_s_x_x}
\end{equation}
and 
\begin{equation}
r_{\phi}(x) \equiv p_{\mathcal D}(x),
\end{equation}
where $p_{\mathcal D}(x)$ is the underlying distribution of individual observations. The joint distribution of the observations and style representations of other group members is $r_{\phi}(x,s)=r_{\phi}(x)r_{\phi}(s | x )$, furthermore, it can be easily shown that $r_{\phi}(s)=q_\phi(s)$ (see the supplementary material). Moreover let us denote the factorized distribution $r_{\phi}(x)r_{\phi}(s)$ by ${\bar r}_{\phi}(x,s)$. We can express the mutual information between data samples and the style representations of other group members as the mutual information between $s$ and $x$ w.r.t. the distribution $r_{\phi}$:
\begin{align}
I_{r_{\phi}}(x;s)&=KL(r_{\phi}(x,s) \| {\bar r}_{\phi}(x,s)) \nonumber \\
&=\mathbb{E}_{r_{\phi}(x,s)} \hspace{2px} \log\frac{r_{\phi}(x,s)}{{\bar r}_{\phi}(x,s)}.
\label{eq:mutualInformation}
\end{align}
The goal of the proposed method is to maximize the lower bound in \eq{eq:fullELBO}, while keeping $I_{r_{\phi}}(x; s)$ low enough to make $s$ invariant to content and thus force the encoder to use only the latent vector $c$ to represent content attributes. Instead of handling the problem as a constrained optimization task, we propose to take its Lagrangian relaxation by penalizing the mutual information, which leads to the following training objective:
\begin{equation}
\begin{aligned}
\underset{\theta,\phi}{\text{maximize}} & & {\mathcal L}_{\theta,\phi}({\mathcal D}) - \lambda I_{r_{\phi}}(x; s),
\label{eq:objective}
\end{aligned}
\end{equation}
where the parameter $\lambda$ controls the weight of the mutual information term. We can estimate the mutual information term in \eq{eq:objective} using a parametric neural estimator~\cite{Belghazi2018} as it is possible to generate samples from both $r_{\phi}(x,s)$ and ${\bar r}_{\phi}(x,s)$. For example, to generate a sample from $r_{\phi}(x,s)$ we can first choose a group ${\bf x}^n$ and pick two members $x^n_i$ and $x^n_j$ ($i\neq j$) uniformly at random. Then let $x=x^n_i$ and $s\sim q_\phi(s| x^n_j)$. The samples from ${\bar r}_{\phi}(x,s)$ are pairs of independent points from $r_{\phi}(x)=p_{\mathcal D}(x)$ and $r_{\phi}(s)=q_\phi(s)$. Thus to generate a sample, we can first pick two observations $x^n_i\sim p_{\mathcal D}(x)$ and $x^m_j \sim p_{\mathcal D}(x)$, then choose $x=x^n_i$ and $s\sim q_\phi(s| x^m_j)$.

As usually in practice, we train our models using mini-batch based stochastic optimization methods, thus it is the most efficient to draw samples from $r_{\phi}(x,s)$ and ${\bar r}_{\phi}(x,s)$ based on the same mini-batches of groups that we use to maximize \eq{eq:fullELBO}. The pseudo-code of our mini-batch based sampling approach is described in \alg{al:sampling}. We note that to generate a sample from ${\bar r}_{\phi}(x,s)$ we never choose $x$ and $s$ from the same group otherwise such pairs would be over-represented. For the same reason, to generate samples from $r_{\phi}(x,s)$ a given data point $x$ is never paired with a style vector sampled from its style posterior $q_\phi(s|x)$. 

\LinesNumbered
\begin{algorithm}[t]
\SetInd{2.0em}{0.0em}
\DontPrintSemicolon
\caption{Generating samples from $r_{\phi}(x,s)$ and ${\bar r}_{\phi}(x,s)$ \label{al:sampling}}
\SetKwComment{Comment}{      \#\ }{}
\Input{$B=\{ {\bf x}^b : b=1,\dots,N_B\}$.}
\BlankLine
$R \gets \emptyset$, ${\bar R}\gets \emptyset$.\;
\ForEach{${\bf x}^b\in B $ } {
  Choose $i$ and $j$ ($i\neq j$) uniformly at random from $\{1,\dots,K\}$.\;
  Sample $s^b_j\sim q_\phi(s|x^b_j)$.\;
  $R \gets R \cup \{(x^b_i, s^b_j)\}$.\;
  $m \gets 1+(b \mod N_B)$ \Comment*[l]{Index of next group}
  Choose $k$ uniformly at random from $\{1,\dots,K\}$.\;
  Choose $l$ uniformly at random from $\{1,\dots,K\}$.\;
  Sample $s^m_l\sim q_\phi(s|x^m_l)$.\;
  ${\bar R} \gets {\bar R} \cup \{(x^b_k, s^m_l)\}$.\;
}
\Return{$R$, ${\bar R}$}
\end{algorithm}

Since samples from $r_{\phi}(x,s)$ and ${\bar r}_{\phi}(x,s)$ are accessible, it is possible to estimate $I_{r_{\phi}}( x; s )$ as:
\begin{equation}
 I_{r_{\phi}}( x; s ) \geq \sup_\psi\big\{\mathbb{E}_{r}[T_\psi(x, s)]-\log(\mathbb{E}_{\bar r}[e^{T_\psi(x, s)}]) \big\},
\end{equation}
where $T_\psi$ is a neural network parameterized by $\psi$, see ~\cite{Belghazi2018,Nguyen2010}. Let $R$ and ${\bar R}$ be sets of samples from $r_{\phi}(x,s)$ and ${\bar r}_{\phi}(x,s)$, respectively. We train $T_\psi$ to maximize
\begin{equation}
{\mathcal L}_{\psi}(R, {\bar R}) = \sum_{(x,s)\in R}\frac{T_\psi(x, s)}{|R|}-\log\sum_{(x,s)\in{\bar R}}\frac{e^{T_\psi(x, s)}}{|{\bar R}|},
\end{equation}
and use this approximation to minimize the mutual information in \eq{eq:objective} in an adversarial manner. To achieve a sufficiently small mutual information, we adaptively adjust the weight of the mutual information term in \eq{eq:objective} in each iteration of the training process:
\begin{equation}
\lambda \gets \lambda + \alpha \big({\mathcal L}_{\psi}(R, {\bar R})/I^\star-1\big),
\label{eq:lambdaUpdate}
\end{equation}
where $I^\star$ is the target value for the mutual information (we set $I^\star=0.2$ in our experiments), ${\mathcal L}_{\psi}(R, {\bar R})$ is evaluated based on the current training mini-batch, while $\alpha$ is a step-size (we used $\alpha=0.1$ in our experiments). This update rule increases (decreases) the weight $\lambda$ if the current estimation of the mutual information is higher (lower) than the target value $I^\star$.

\LinesNumbered
\begin{algorithm}[t]
\SetInd{2.0em}{0.0em}
\DontPrintSemicolon
\caption{MLVAE with Adversarial Disentanglement\label{al:mlvaead}}
\Input{Dataset $\{ {\bf x}^n : n=1,\dots,N\}$, mini-batch size $N_B$, number of training iterations $it$, number of $T_\psi$ update steps $it_T$, optimizers $g_{\theta,\phi}$ and $g_\psi$.}
\BlankLine
Initialize $\theta$,$\phi$, $\psi$, and $\lambda$.\;
\RepTimes{$it$}{
Sample a mini-batch of observation groups $B=\{ {\bf x}^b : b=1,\dots,N_B\}$.\;
Use \alg{al:sampling} with $B$ as input to obtain two sets $R$ and ${\bar R}$ containing samples from $r_{\phi}(x,s)$ and ${\bar r}_{\phi}(x,s)$, respectively.\;
$\theta,\phi\!\gets\! g_{\theta,\phi} \big(\!-\!{\mathcal L}_{\theta,\phi}(B) + \lambda{\mathcal L}_{\psi}(R, {\bar R})\big)$.\;
Update $\lambda$ using \eq{eq:lambdaUpdate}.\;
\RepTimes{$it_T$}{
Sample a mini-batch of groups $B=\{ {\bf x}^b : b=1,\dots,N_B\}$.\;
Use \alg{al:sampling} with $B$ as input to obtain two sets $R$ and ${\bar R}$ containing samples from $r_{\phi}(x,s)$ and ${\bar r}_{\phi}(x,s)$, respectively.\;
$\psi\! \gets \! g_{\psi} \big( -{\mathcal L}_{\psi}(R, {\bar R}) \big)$.\;
}
}
\end{algorithm}

As experimental results prove, the method described in this section makes the style variable $s$ less informative about the content, thus forcing the method to learn more effective content representation in $c$. We refer to MLVAE improved with our adversarial disentanglement method as MLVAE-AD. The pseudo code in~\alg{al:mlvaead} shows the main steps of the proposed algorithm.

\begin{figure*}[t]
\subfigcapskip 1mm
\centering
\setlength{\tabcolsep}{1pt}
\begin{tabular}{c@{\hskip 0.5cm}c}
\subfigure[MLVAE]{
\begin{tabular}{cc}
\swapped{0.192\linewidth}{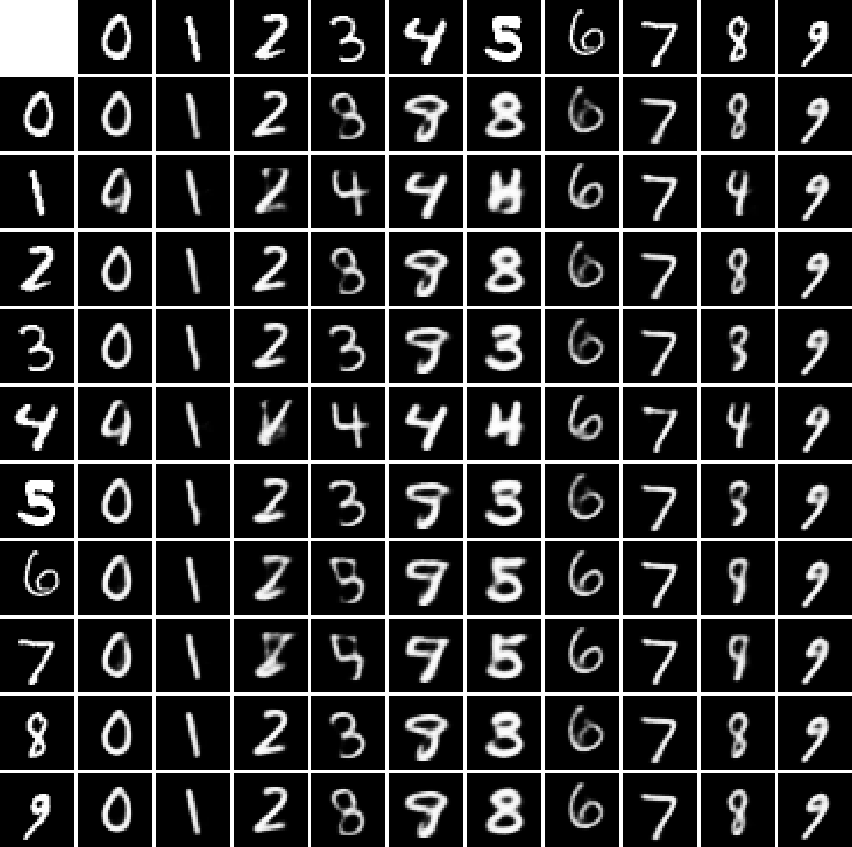} &
\raisebox{-.5\height}{\includegraphics[height=0.192\linewidth]{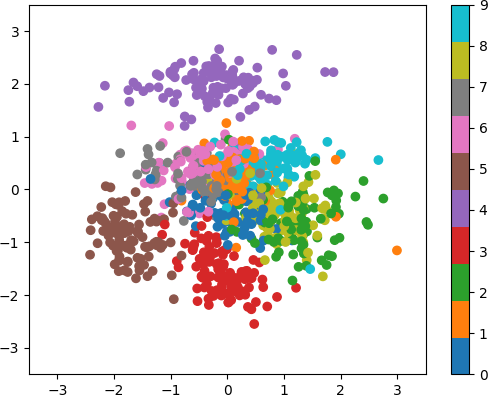}}
\end{tabular}
} &
\subfigure[MLVAE-AD]{
\begin{tabular}{cc}
\swapped{0.192\linewidth}{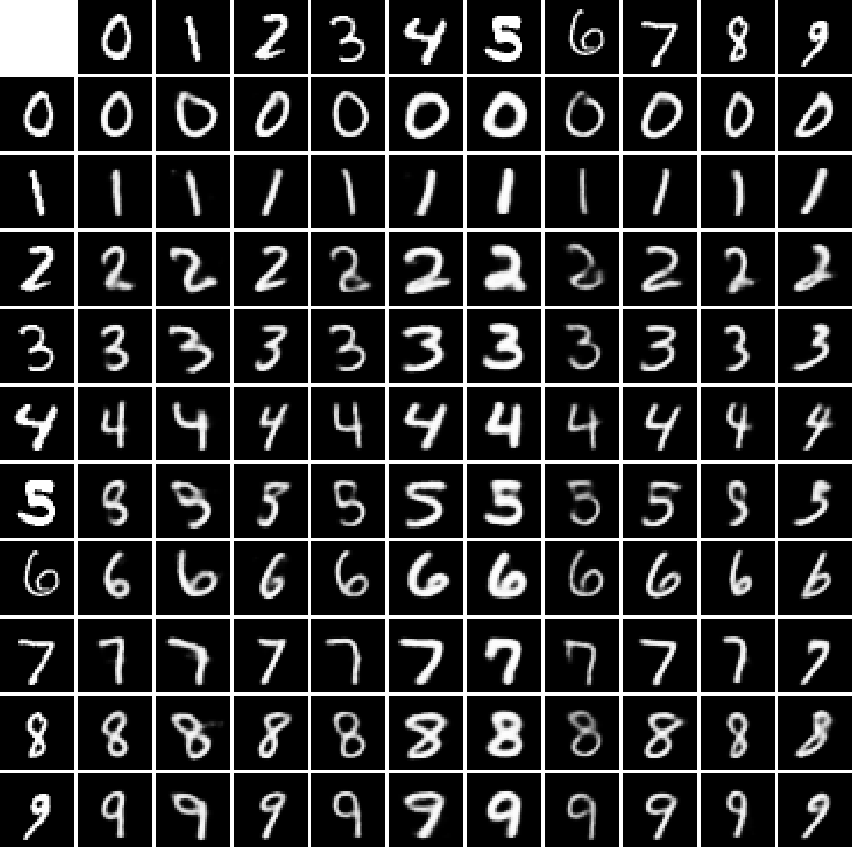} &
\raisebox{-.5\height}{\includegraphics[height=0.192\linewidth]{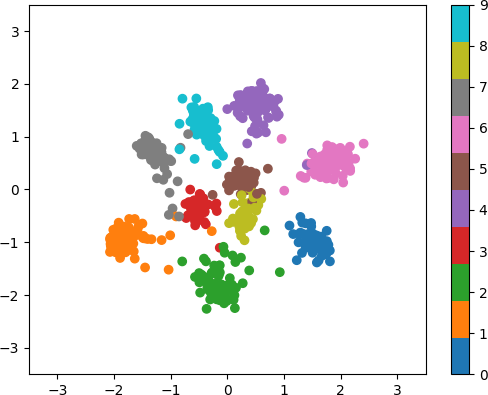}}
\end{tabular}
}
\end{tabular}
\caption[]{Example results from the qualitative experiments on MNIST using $d_c=2$, $d_s=14$, and $K=2$. The figure contains generated images obtained from swapped encodings (see text) and scatter plots for 2-dimensional content embeddings.}
\label{fig:mnist}
\end{figure*}

\section{Experiments}
\label{se:experiments}
In our experiments we focused on testing the effect of the proposed adversarial disentanglement approach. We used three image datasets that exhibit different variability in both content and style. To evaluate the disentanglement in the learned representations simple classifiers were trained on both the content and style mean variables. For that purpose, each of the datasets had been split into three parts. For a given dataset, the first set was used to form the groups for training the models, the classifiers were trained on the second, while the classification performances were evaluated on the last part. Herein, we present the classification results obtained by standard SVM. In the supplemental, evaluation using linear regression classifier shows that such a simpler method can also easily distinguish different classes in the content representations, but performs weaker in discovering content information in style encodings.

\begin{table}[t]
\caption{Evaluation of the results obtained on the MNIST test set using different numbers of latent dimensions and group sizes. Results for $d_c=2,d_s=14$ and $d_c=8,d_s=8$ are shown in rows 1-3 and 4-6, respectively.}
\label{tab:resultsMNIST}
\centering
\begin{tabular}{ccccccc}
\toprule
K & \multicolumn{3}{c}{MLVAE} & \multicolumn{3}{c}{MLVAE-AD}\\
 & ${\mathcal C}(c)$ & ${\mathcal C}(s)$ & ${\mathcal L}_{rec}$ & ${\mathcal C}(c)$ & ${\mathcal C}(s)$ & ${\mathcal L}_{rec}$ \\
\cmidrule(lr){2-4} \cmidrule(lr){5-7}
2 &    65.0\% & 89.2\% & 75.2     &    97.6\% & 41.2\% & 78.2 \\
5 &    92.9\% & 85.2\% & 76.0     &    97.3\% & 40.2\% & 80.6 \\
10 &    94.1\% & 85.9\% & 75.7     &    96.3\% & 57.9\% & 79.5 \\ [1.0ex]
2 &    84.6\% & 79.6\% & 77.4     &    98.4\% & 26.8\% & 79.6 \\
5 &    97.3\% & 61.9\% & 81.5     &    98.5\% & 29.4\% & 85.3 \\
10 &    97.6\% & 60.9\% & 83.5     &    98.0\% & 30.6\% & 86.0 \\
\bottomrule
\end{tabular}
\end{table}

The MNIST~\cite{Lecun1998} dataset is composed of only 10 classes of handwritten digits, while there is a wide range of variability in style. We split the set of 50000 training images into two parts. The models were trained on 45000 samples, in case of a given group size we randomly formed 10000 groups from each of the 10 classes. The remaining 5000 images were used to train the classifiers, while the MNIST test set was used for evaluation. The original $28\times 28$ pixels images were padded to $32\times 32$ pixels to fit to the network architecture.

The Chairs~\cite{Yang2015} dataset contains rendered images of about one thousand different three-dimensional chair models, but the intra-class variability is very low as the images within a given class differ only by the view of the model. The version of this dataset that we used contains $64\times 64$ pixel images of 809 different three-dimensional chair models rendered from 62 different views. In the experiments, our algorithm was trained on 659 randomly chosen classes, while the remaining 150 classes were used for evaluation. For a given group size, 100 random groups were formed from each of the 659 training classes. From each of the 150 test classes we used 31 random images to train the classifiers, while their accuracies were measured on the remaining 31 samples.

Finally, the VGGFace2~\cite{Cao2018} face recognition dataset represents high variability in both content and style. It contains more than 3 million in-the-wild images of about 9000 identities. The training set consist of 8631 classes, for each class we formed 50 random groups for a given group size. From each of the 500 test classes, we randomly selected 50 images to train the classifiers and 10 other images for evaluation. The images were aligned based on facial keypoints provided with the dataset and resized to $64\times 64$ pixels.

\begin{table}[t]
\caption{Results of MLVAE on MNIST ($d_c=2$, $d_s=14$, $K=2$) with increased style regularization weight $\beta$.}
\label{tab:stylereg}
\centering
\begin{tabular}{cccc}
\toprule
$\beta$ & ${\mathcal C}(c)$ & ${\mathcal C}(s)$ & ${\mathcal L}_{rec}$\\
\midrule
$1.5$  &    87.6\% & 77.4\% & 80.4 \\                    
$2.0$  &    93.5\% & 55.5\% & 85.2 \\
$5.0$  &    95.6\% & 30.9\% & 106.8 \\
$10.0$  &    94.3\% & 20.3\% & 128.7 \\
$20.0$  &    94.5\% & 29.2\% & 141.3 \\
\bottomrule
\end{tabular}
\end{table}

For the Chairs and VGGFace2 datasets the classes used to train the models were disjoint from the classes used for evaluation. Thus, experiments on these sets show the ability of the method to generalize to unseen contents. 

\begin{figure*}[t]
\subfigcapskip 1mm
\centering
\setlength{\tabcolsep}{2pt}
\begin{tabular}{cc} 
\subfigure[Chairs\label{fig:resultsChairs}]{\swapped{0.44\linewidth}{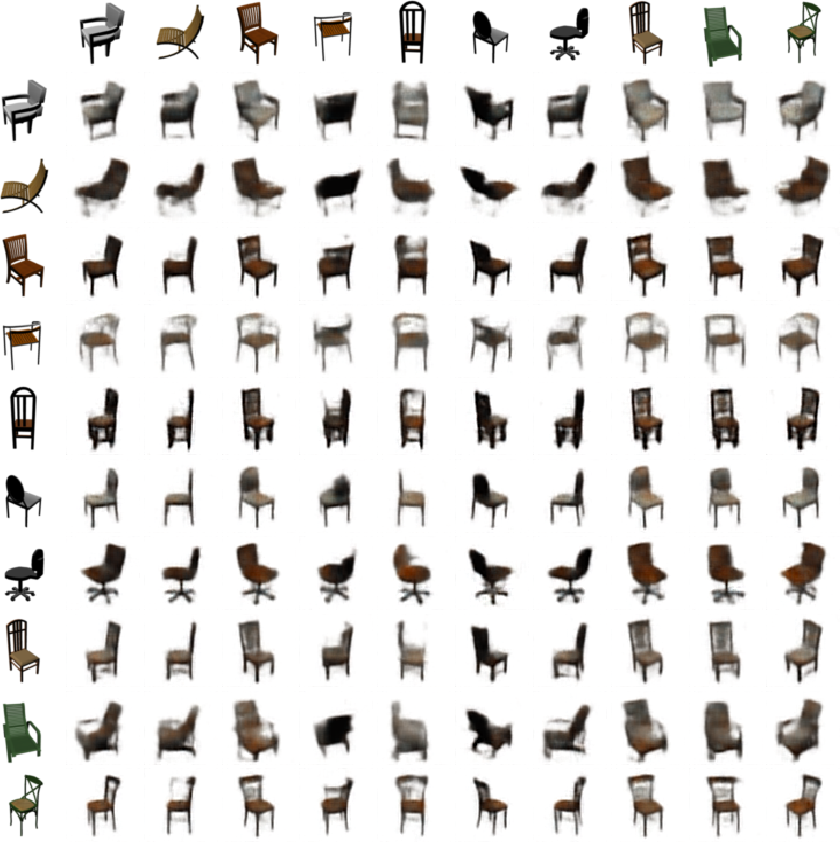}} \hfill & 
\subfigure[VGGFace2\label{fig:resultsVGGFace2}]{\swapped{0.44\linewidth}{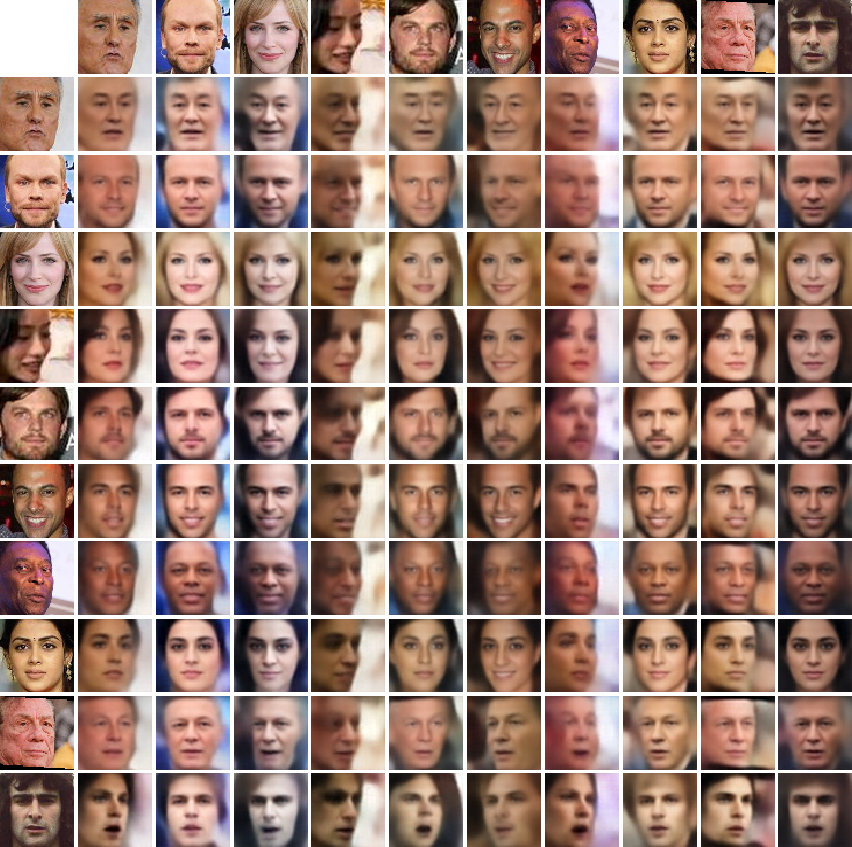}}
\end{tabular}
\caption[]{Qualitative experiments on Chairs ($d_c=32, d_s=32, K=2$) and VGGFace2 ($d_c=64, d_s=64, K=2$) datasets: decoding outputs from swapped representations.}
\end{figure*}

To implement the encoder, the decoder, and $T_\psi$ we used simple convolutional neural networks. Note that $T_\psi$ is unusual in the sense that it takes a (higher dimensional) data point and a (lower dimensional) representation vector as input. To implement such a network, we first used a similar neural-network structure that we used for the encoder to map $x$ into a lower dimensional vector. Then we concatenated the resulting vector with $s$ and applied some more layers to obtain the output of $T_\psi$. Note that the neural network architectures and training hyper-parameters were not fine-tuned to obtain high quality results. Adam optimizer~\cite{Kingma2015} was applied in all of the experiments. The description of the neural networks and the optimization parameters used for the different experiments can be found in the supplementary material.

For qualitative evaluation we created image plots with {\it swapped latents}. For such image grids, we encoded a few images from the test datasets and swapped the content and style vectors to obtain new representations. These representations were then fed into the decoder to generate the images. In such an image grid, the first row and the left-most column contain the input samples.

\begin{table}[t]
\caption{Quantitative results on Chairs test set with $d_c=32$, $d_s=32$, and different group sizes. }
\label{tab:resultsChairs}
\centering
\resizebox{.99\columnwidth}!{
\begin{tabular}{ccccccc}
\toprule
K & \multicolumn{3}{c}{MLVAE} & \multicolumn{3}{c}{MLVAE-AD}\\
& ${\mathcal C}(c)$ & ${\mathcal C}(s)$ & ${\mathcal L}_{rec}$ & ${\mathcal C}(c)$ & ${\mathcal C}(s)$ & ${\mathcal L}_{rec}$ \\
\cmidrule(lr){2-4}
\cmidrule(lr){5-7}
2  & 76.3\% & 74.5\% & 85.7     &    93.8\% &  8.6\% & 111.7 \\
5   & 82.4\% & 72.3\% & 89.8     &    95.0\% & 15.4\% & 110.7 \\
10   & 92.3\% & 60.9\% & 97.9     &    92.8\% & 25.7\% & 113.1 \\
\bottomrule
\end{tabular}
}
\end{table}

\subsection{Evaluation}
When the representation is successfully disentangled, we expect to obtain high classification accuracy on content representations while low accuracy on style representations. In our tables, the classification accuracies obtained on the content and style representations are denoted by ${\mathcal C}(c)$ and ${\mathcal C}(s)$, respectively. We also present the average reconstruction errors (average per-sample negative log-likelihoods) obtained on the test sets which is denoted by ${\mathcal L}_{rec}$. A significant increase in reconstruction error would suggest that adversarial disentanglement weakens the representation capacity of the model. In our experiments, we focused on small group sizes ($K=2,5,10$). For a given setting, the models were trained both with and without adversarial disentanglement. 

In case of the MNIST handwritten digits, the focus of the experiments was on examining whether the proposed method is capable of forcing the model to represent the 10 classes on a very low-dimensional content vector ($d_c=2$). We tested the algorithm for $d_c=2, d_s=14$ and for $d_c=8, d_s=8$ as well. Results in \tabl{tab:resultsMNIST} show that the original MLVAE method obtains poor disentanglement when the group size is small. At the same time when adversarial training is applied, the method can separate the content and style attributes well. \fig{fig:mnist} shows an example where MLVAE failed to use only $c$ to represent the content, while MLVAE-AD formed a content distribution in which the 10 classes are well separated. Additional experiments on the MNIST dataset can be found in the supplementary material.

\begin{table}[t]
\caption{Results on the test set of VGGFace2 with $d_c=64$, $d_s=64$, and different group sizes.}
\label{tab:resultsVGGFace2}
\centering
\resizebox{.99\columnwidth}!{
\begin{tabular}{ccccccc}
\toprule
K & \multicolumn{3}{c}{MLVAE} & \multicolumn{3}{c}{MLVAE-AD}\\
& ${\mathcal C}(c)$ & ${\mathcal C}(s)$ & ${\mathcal L}_{rec}$ & ${\mathcal C}(c)$ & ${\mathcal C}(s)$ & ${\mathcal L}_{rec}$ \\
\cmidrule(lr){2-4}
\cmidrule(lr){5-7}
2  & 21.4\% & 27.5\% & 209.4     &    39.0\% &  6.8\% & 260.9 \\
5   & 31.7\% & 27.9\% & 212.7     &    36.8\% & 13.1\% & 251.0 \\
10  & 39.6\% & 27.5\% & 215.5     &    41.1\% & 14.1\% & 249.9 \\
\bottomrule
\end{tabular}
}
\end{table}

As it was mentioned previously, one might expect that the original MLVAE method could obtain better disentanglement with increased regularization weight on the style posterior, as it would force the model to use the content vector. This would be a simple alternative to our adversarial approach. On the other hand, increased regularization decreases the mutual information between the latents and the data, thus it suppresses not only the content but also the style information in $s$. To show this, we assigned a weight $\beta$ to the second term of the MLVAE objective:
\begin{align}
&\mathbb{E}_{q_\phi(c, {\bf s} | {\bf x})}  \sum_{i=1}^K\log p_\theta(x_i | c, s_i ) \nonumber\\
&-\beta \sum_{i=1}^K KL(q_\phi(s_i|x_i)|p(s_i)) - KL(q_\phi(c|{\bf x})|p(c)).
\label{eq:betaGroupELBO}
\end{align}
Results for different values of $\beta$ can be found in~\tabl{tab:stylereg}. The table shows classification accuracies and average reconstruction errors measured on the MNIST test set. 
Comparing these results to the first row of \tabl{tab:resultsMNIST}, it can be seen that although increased regularization results in better disentanglement, it also weakens the overall capacity of the model (higher reconstruction errors) while the content representations are less effective compared to those of the proposed method.

In case of the Chairs dataset, the intra-class variability is restricted to the view (rotation) of the objects and this low style variability can be efficiently represented with a low dimensional style variable, \eg, $d_s=2$~\cite{Hosoya2019}. Therefore, we tested the performance of our adversarial approach for a case when the style variable is strongly over-parameterized ($d_s=32$). Experimental results presented in \tabl{tab:resultsChairs} and \fig{fig:resultsChairs} show that the proposed method efficiently eliminates content related information from $s$.

\begin{figure}[t]
\centering
\resizebox{.99\columnwidth}!{
\swapped{0.99\linewidth}{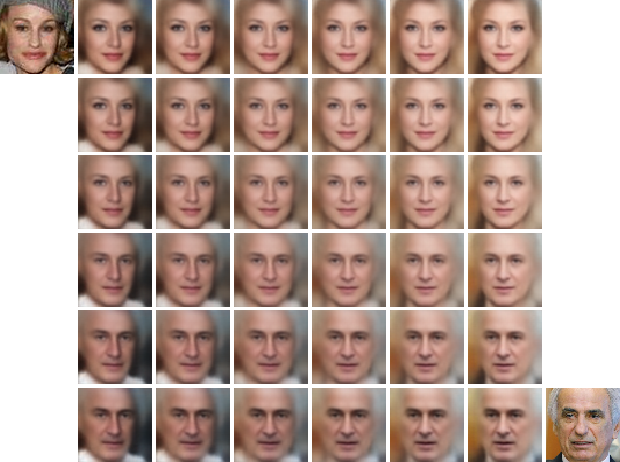}
}
\caption{Latent traversals between encodings of test images at the upper-left and bottom-right. These results are from the experiment with $K=2$.\label{fig:traversalReal}}
\end{figure}

Separating the facial identity from other factors like facial expression, pose or illumination is a more challenging task.  Quantitative results in \tabl{tab:resultsVGGFace2} shows that adversarial disentanglement improves the results of the original MLVAE method while \fig{fig:resultsVGGFace2} and \fig{fig:traversalReal} demonstrate that the proposed method was able to separate the most important attributes corresponding to facial identity. Additional qualitative experiments can be found in the supplementary material.

\subsection{Comparisons}
To learn representations that are invariant to specific factors, an  autoencoder based approach was recently proposed by \citeauthor{Jaiswal2018}~(\citeyear{Jaiswal2018}). This method is similar to the current work as it is unsupervised on the considered nuisance factors and applies adversarial training to induce invariance, but differs in that it requires supervision on the target variables and in its non-probabilistic nature. We considered their experiment on a rotated version of MNIST (MNIST-ROT), using the same single hidden layer network architectures, except that our encoder outputs latent variances as well. Both the method and the classifier were trained on $\Theta\in\{0^{\circ}, \pm 22.5^{\circ}, \pm 45^{\circ}\}$ with $d_c=10, d_s=20$, $K=2$ and tested with rotations $\Theta$, $\pm 55^{\circ}$, and $\pm 65^{\circ}$. The results and comparison to \cite{Jaiswal2018} can be found in~\tabl{tab:resultsMNISTROT}.

As a further experiment, we considered \cite{Kingma2014b}. Although this method was proposed for semi-supervised learning, it can also be used in a supervised setting. For the MNIST dataset with all available training labels a test-set accuracy of $99.04\%$ was reported. For comparison we employed the same two-hidden MLPs and set $d_c=10$ and $d_s=50$. Averaged over 10 runs, using our method we achieved a classification accuracy of $97.86\%\;(\pm 0.13)$, while with the simplified content accumulation strategy of \cite{Hosoya2019} using our method we obtained $98.09\%\;(\pm 0.08)$. 

These results indicate that our adversarial method renders grouped observations based learning competitive to supervised methods that learn from explicit target labels. 

\begin{table}[t]
\caption{Classification results (${\mathcal C}(c)$) and comparison on MNIST-ROT. For the MLVAE and MLVAE-AD methods, means and standard deviations over 10 runs are presented.}
\label{tab:resultsMNISTROT}
\centering
\resizebox{.99\columnwidth}!{
\begin{tabular}{cccc}
\toprule
& \citeauthor{Jaiswal2018} & MLVAE & MLVAE-AD \\
\cmidrule(lr){2-2}
\cmidrule(lr){3-3}
\cmidrule(lr){4-4}
$\Theta$ &  {\bf 97.7\%} & 52.2\% \small{($\pm$ 3.8)} & 95.1\% \small{($\pm$ 0.2)} \\
$\pm 55^{\circ}$ & 85.6\% & 41.9\%  \small{($\pm$ 3.4)} & {\bf 85.7\%} \small{($\pm$ 0.6)} \\
$\pm 65^{\circ}$ & 69.6\% & 29.3\% \small{($\pm$ 3.2)} & {\bf 71.0\%} \small{($\pm$ 0.9)} \\
\bottomrule
\end{tabular}
}
\end{table}

\section{Conclusion}
This paper proposed an adversarial disentanglement approach to improve the results of grouped observations based methods. To eliminate content related information in the style variable, the algorithm minimizes the predictability of the style representation of a given data sample from other members within the same group. This idea was formulated as the minimization of an appropriately constructed mutual information term. We proposed to estimate this mutual information using a parametric neural estimator and train the encoder in an adversarial manner. Experiments and comparisons on image datasets demonstrated the efficacy of the proposed method in separating the content and style related attributes and its ability to generalize to unseen classes.

\bibliographystyle{aaai}
\fontsize{9.5pt}{10.5pt} \selectfont
\bibliography{aaai20}
\end{document}